\documentclass[12pt]{article}
\usepackage{amsmath,amsfonts,amssymb,amsthm,amsbsy}
\usepackage{amsfonts}
\usepackage{amssymb}
\usepackage{epsfig}
\usepackage{color,url}
\usepackage[round]{natbib}
\usepackage{subfigure}
\DeclareMathSizes{12}{12}{10}{10} 
\definecolor{Red}{rgb}{1,0,0}

\DeclareMathOperator*{\argmin}{\arg\!\min}

\newcommand{\BEAS}{\begin{eqnarray*}}
\newcommand{\EEAS}{\end{eqnarray*}}
\newcommand{\BEA}{\begin{eqnarray}}
\newcommand{\EEA}{\end{eqnarray}}
\newcommand{\BIT}{\begin{itemize}}
\newcommand{\EIT}{\end{itemize}}
\newcommand{\BNUM}{\begin{enumerate}}
\newcommand{\ENUM}{\end{enumerate}}

\newtheorem {Thm}{Theorem}
\newtheorem {prop} {Proposition}

\newcommand{\bp}{\begin{prop}}
\newcommand{\ep}{\end{prop}}

\def \beq {\begin{equation} }
\def \eeq {\end{equation} }

\def \bea {\begin{eqnarray} }
\def \eea { \end{eqnarray} }
\def \beas {\begin{eqnarray*} }
\def \eeas { \end{eqnarray*} }

\def \l({\left(}
\def \r){\right)}

\def \bi {\begin{itemize}}
\def \ei {\end{itemize}}

\begin{document}
\title{Regression shrinkage and grouping of highly correlated predictors with HORSES}

\author{
Woncheol Jang \\ University of Georgia
\footnote{Email:jang@uga.edu} \and Johan Lim
\\ Seoul National University \footnote{Email:johanlim@snu.ac.kr}
\and Nicole A. Lazar \\ University of Georgia\footnote{Email: nlazar@stat.uga.edu} 
\and Ji Meng Loh \\
AT\&T Labs-Research \footnote{Email:loh@research.att.com}
 \and
Donghyeon Yu\\ Seoul National University
\footnote{Email:bunguji@snu.ac.kr} }

\date{\today}
\maketitle
\vspace*{-0.3in}
\begin{abstract}
Identifying homogeneous subgroups of variables can be challenging in
high dimensional data analysis with highly correlated predictors. We propose a new method called Hexagonal Operator for Regression with Shrinkage and Equality Selection, HORSES for short, that simultaneously selects positively correlated variables and identifies them as predictive
clusters. This is achieved via a constrained
least-squares problem with regularization that consists of a linear
combination of an $L_1$ penalty for the coefficients and another $L_1$
penalty for pairwise differences of the coefficients. This specification of the penalty function encourages grouping of positively correlated predictors combined with a sparsity solution. We construct an efficient algorithm to implement the HORSES procedure. We show via simulation that the proposed method outperforms other variable selection methods in terms of prediction
error and parsimony.  The technique is demonstrated on two data  sets, a small data set from analysis of soil in Appalachia, and a high dimensional data set   from a near infrared   (NIR) spectroscopy study, showing the 
flexibility of the methodology.
\vspace{.2in}


\end{abstract}
  
\section{Introduction} \label{sect:intro}
Suppose that we observe $(x_1, y_1), \ldots, (x_n, y_n)$ where $x_i=(x_{i1}, \ldots, x_{ip})^t$ is a $p$-dimensional predictor and $y_i$ is the response variable.  We  consider  a standard linear model for each of $n$ observations
$$
y _i= \sum_{j=1}^p \beta_j x_{ij} +\epsilon_i, \mbox{ for } i =1, \ldots, n,
$$
with E$(\epsilon_i)=0$ and Var$(\epsilon_i) =\sigma^2$.  We also
assume the predictors are standardized and the response variable is
centered,
$$
\sum_{i=1}^n y_i =0, \quad \sum_{i=1}^n x_{ij} = 0  \mbox{ and }
\sum_{i=1}^n x_{ij}^2 =1 \quad  \mbox{      for   } j =1, \ldots, p.
$$

With the dramatic increase in the amount of data collected in many fields comes a corresponding increase in the number of predictors $p$
available in data analyses. For simpler interpretation of the
underlying processes generating the data, it is often desired to have
a relatively parsimonious model. It is often a challenge to
identify important predictors out of the many that are available. This becomes more so when the predictors are correlated.


As a motivating example, consider a study involving near infrared (NIR)  spectroscopy data measurements of cookie dough \citep{Osborne1984}.  Near infrared reflectance spectral measurements were made at 700 wavelengths from 1100 to 2498 nanometers (nm) in steps of 2nm for each of 72 cookie doughs made with a standard recipe. The study aims to predict dough chemical composition using the spectral characteristics of NIR reflectance wavelength measurements.  Here, the number of wavelengths $p$ is much bigger than the sample size $n$.

Many methods have been developed to address this issue of high dimensionality. Section \ref{sect:review} contains a brief review. Most of these methods involve minimizing an objective function, like the negative log-likelihood, subject to certain constraints, and the methods in Section \ref{sect:review} mainly differ in the constraints used.

In this paper, we propose a  variable selection procedure that can cluster predictors using the positive correlation structure and is also applicable to data with $p >n$. The constraints we use balance between an $L_1$ norm of the coefficients and an $L_1$ norm for pairwise differences of the coefficients.
We call this procedure a {\em Hexagonal
Operator for Regression with Shrinkage and Equality Selection}, 
HORSES for short, because the constraint region can be represented by a hexagon. The hexagonal shape of the constraint region focuses selection of groups of predictors that are positively correlated.

The goal is to obtain a
homogeneous subgroup structure within the high dimensional predictor
space. This grouping is done by focusing on spatial and/or positive
correlation in the predictors, similar to supervised clustering.
 The benefits of our procedure are a
combination of variance reduction and higher predictive power.

The remainder of the paper is organized as follows.  We introduce the HORSES procedure and its geometric interpretation in Section \ref{sect:model}. We provide an overview of some other methods in Section \ref{sect:review}, relating our procedure with some of these  methods. In Section \ref{sect:compute} we describe the computational
algorithm that we constructed to apply HORSES to data and address the issue of selection of the tuning parameters. A simulation study is
presented in Section \ref{sect:simstudy}.  Two data analyses using HORSES  are
presented in Section \ref{sect:analysis}. We conclude the paper with discussion in
Section \ref{sect:conclusion}. 

\section{Model} \label{sect:model}

In this section we describe our method for variable selection for regression
with {\it positively} correlated predictors. 
Our penalty terms involve a linear combination of an $L_1$
penalty for the coefficients and another $L_1$ penalty for pairwise
differences of coefficients. Computation is done by solving a
constrained least-squares problem.
Specifically, estimates for the HORSES procedure are obtained by solving
\begin{eqnarray}
 \hat \beta &=& \argmin_{\beta} \|y  - \sum_{j=1}^p \beta_j x_j \|^2   \mbox { subject to} \nonumber \\
& & \alpha \sum_{j=1}^p | \beta_j| + (1-\alpha)\sum_{j < k} | \beta_j - \beta_k| \le t, \label{eqn:horses}
\end{eqnarray}
with $d^{-1} \le \alpha \le 1 $ and $d$ is a thresholding
parameter.

\begin{figure}[t]
\centering
\subfigure[Elastic Net]{
\includegraphics[scale=0.25,angle=0]{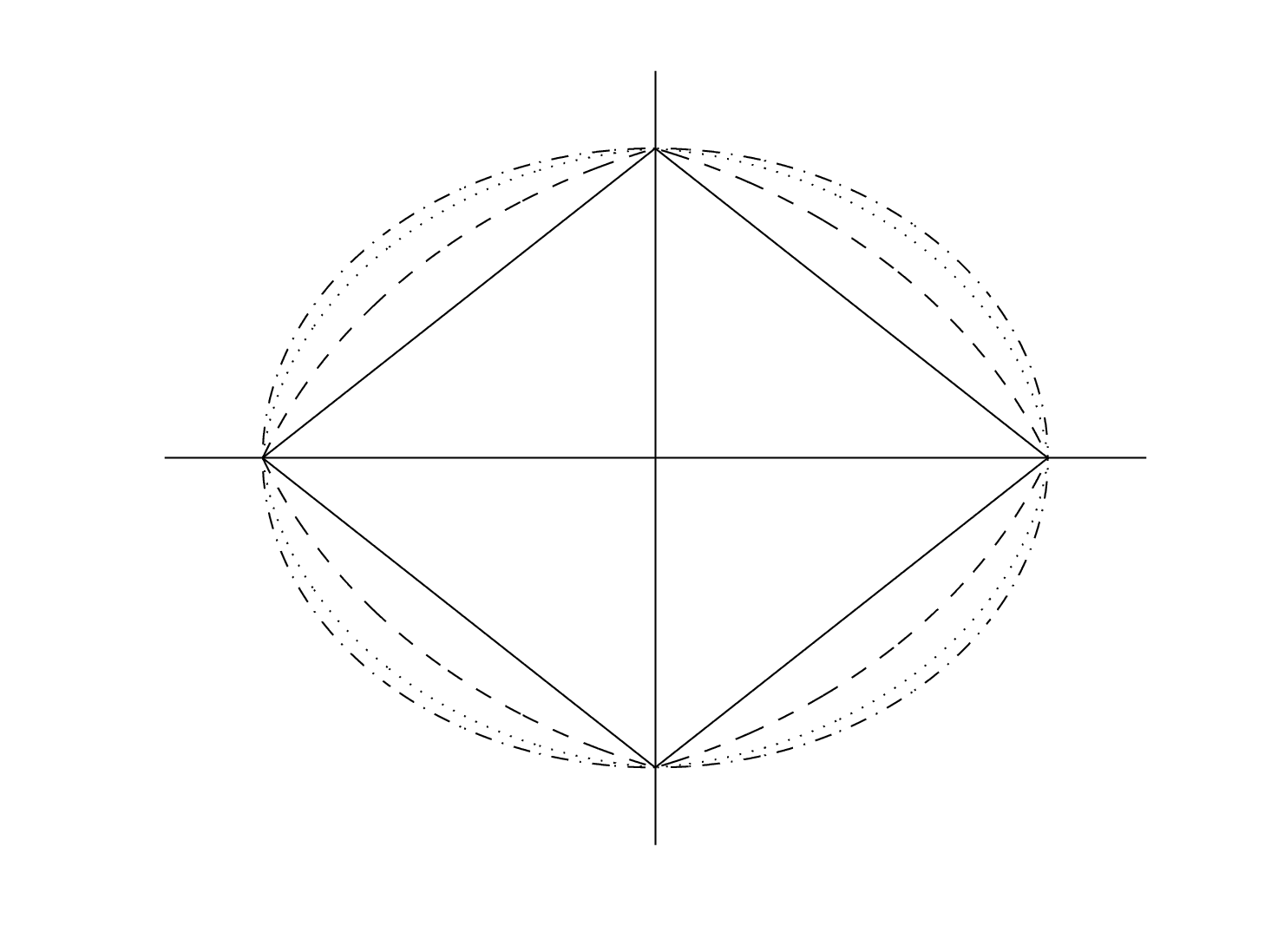}
\label{fig:elastic}
}
 \subfigure[OSCAR]{
\includegraphics[scale=0.25, angle=0]{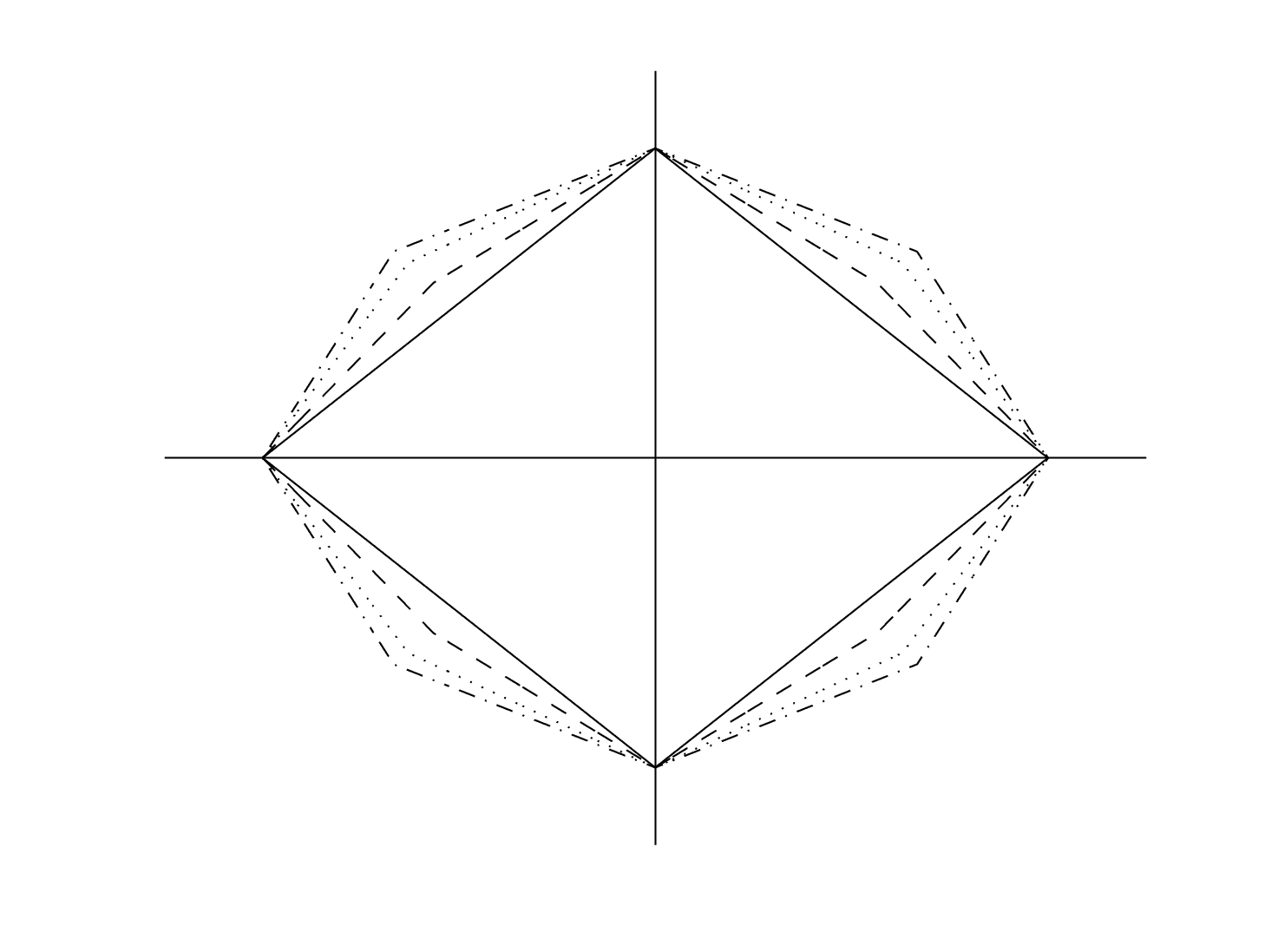}
\label{fig:oscar1}
}
\subfigure[HORSES]{
\includegraphics[scale=0.25, angle=0]{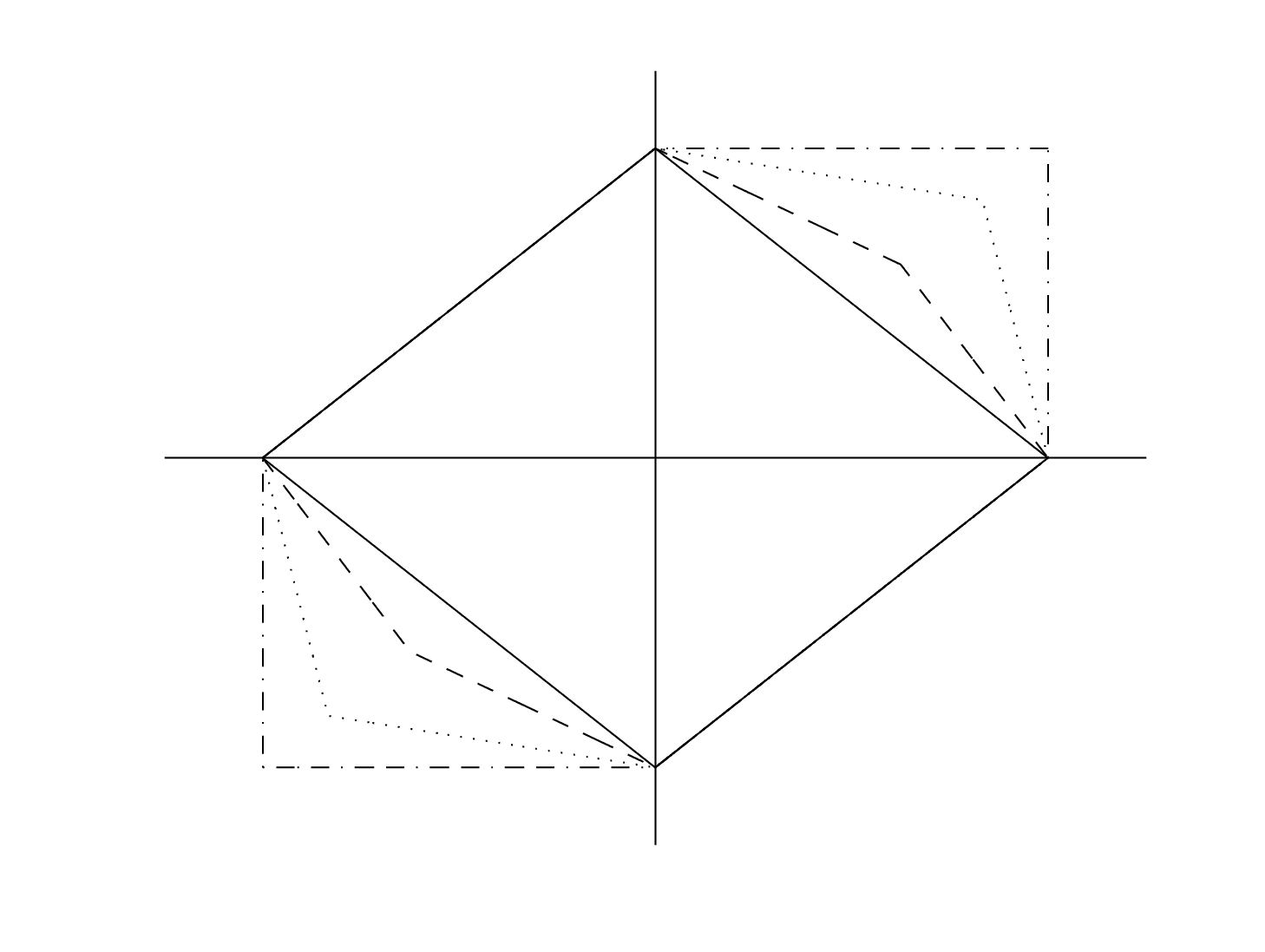}
\label{fig:horses1} } \caption{Graphical representation of the
constraint region in the $(\beta_1, \beta_2)$ plane for
\subref{fig:elastic} Elastic Net, \subref{fig:oscar1} OSCAR, and
\subref{fig:horses1} HORSES}.
\label{fig:region1}
\end{figure}

As we describe in Section \ref{sect:review}, some methods like Elastic Net and OSCAR can group correlated predictors, but they can also put negatively correlated predictors into the same group.  
Our method's novelty is its grouping of
{\it positively} correlated predictors in addition to achieving a sparsity solution. Figure \ref{fig:region1}(c) shows the hexagonal shape of the constraint region induced by (\ref{eqn:horses}), showing schematically the tendency of the procedure to equalize coefficients only in the direction of $y=x$.

The lower bound $d^{-1}$ of $\alpha$ prevents the estimates from being a solution only via the second penalty function, so the HORSES method always achieves sparsity.  We recommend $d=\sqrt{p}$, where $p$ is the number of predictors.
This ensures that the constraint parameter region lies between that of the $L_1$ norm
and of the Elastic Net method, i.e.\ the set of possible estimates for
the HORSES procedure is a subset of that of Elastic Net. In other words, HORSES accounts for positive correlations up to the level of Elastic Net.  With $d=p$, the HORSES parameter
region lies within that of the OSCAR method.

\begin{figure}[t]
\label{fig:likelihood}
\centering
\subfigure[]{
\includegraphics[width=1in,angle=0]{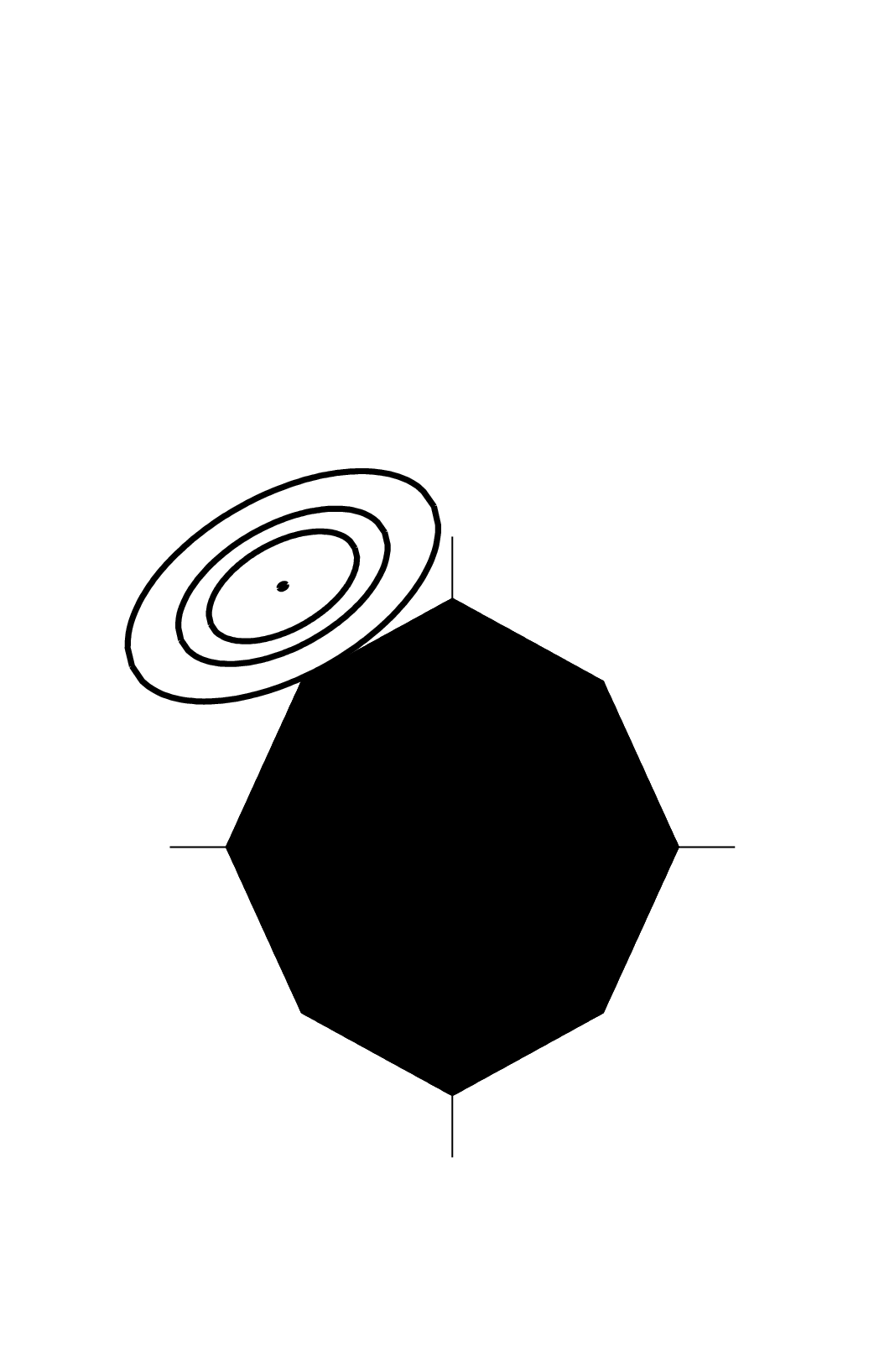}
\label{fig:oscar2}
}
 \subfigure[]{
\includegraphics[width=1in, angle=0]{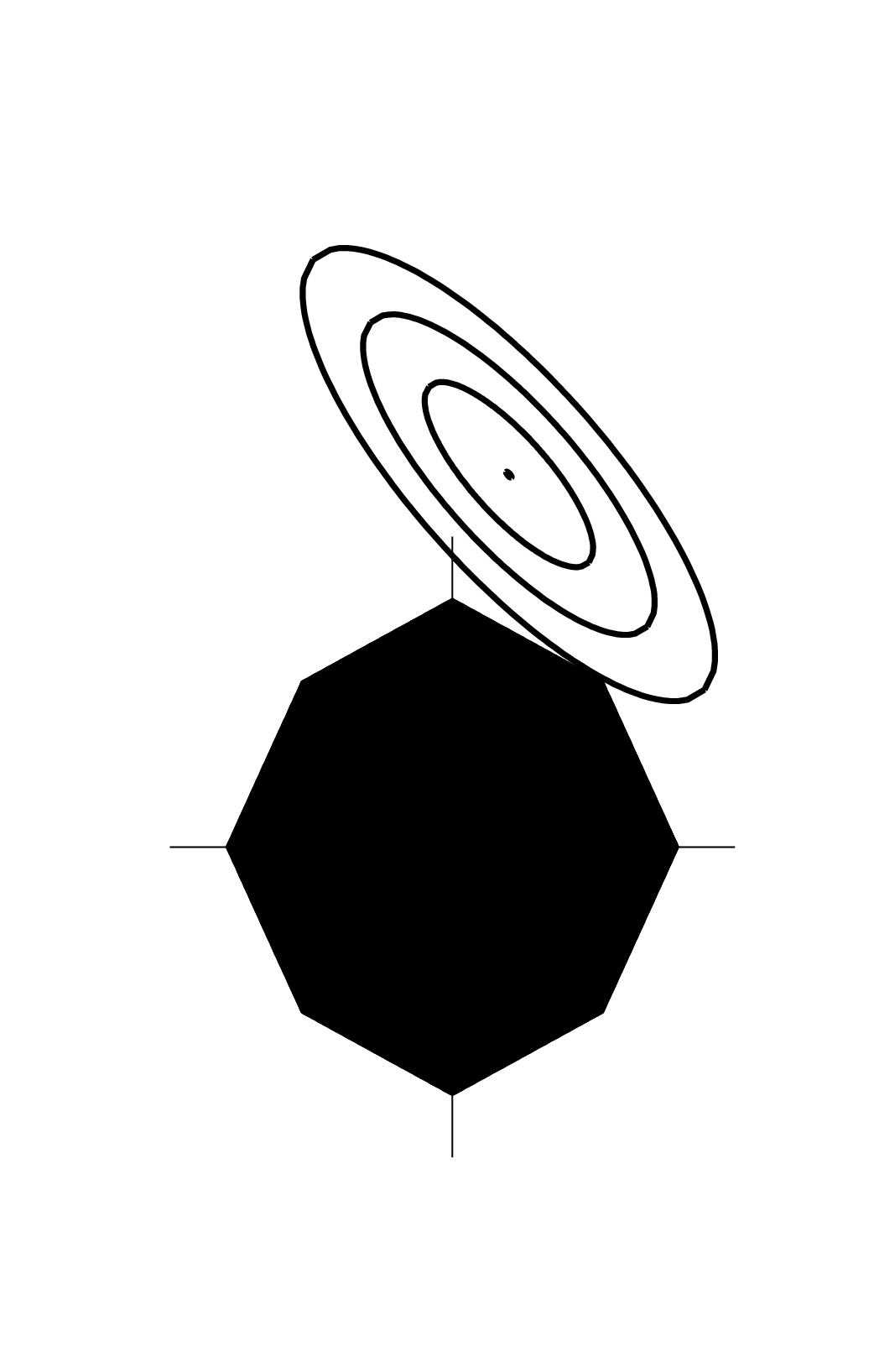}
\label{fig:oscar3}
}
\subfigure[]{
\includegraphics[width=1in,angle=0]{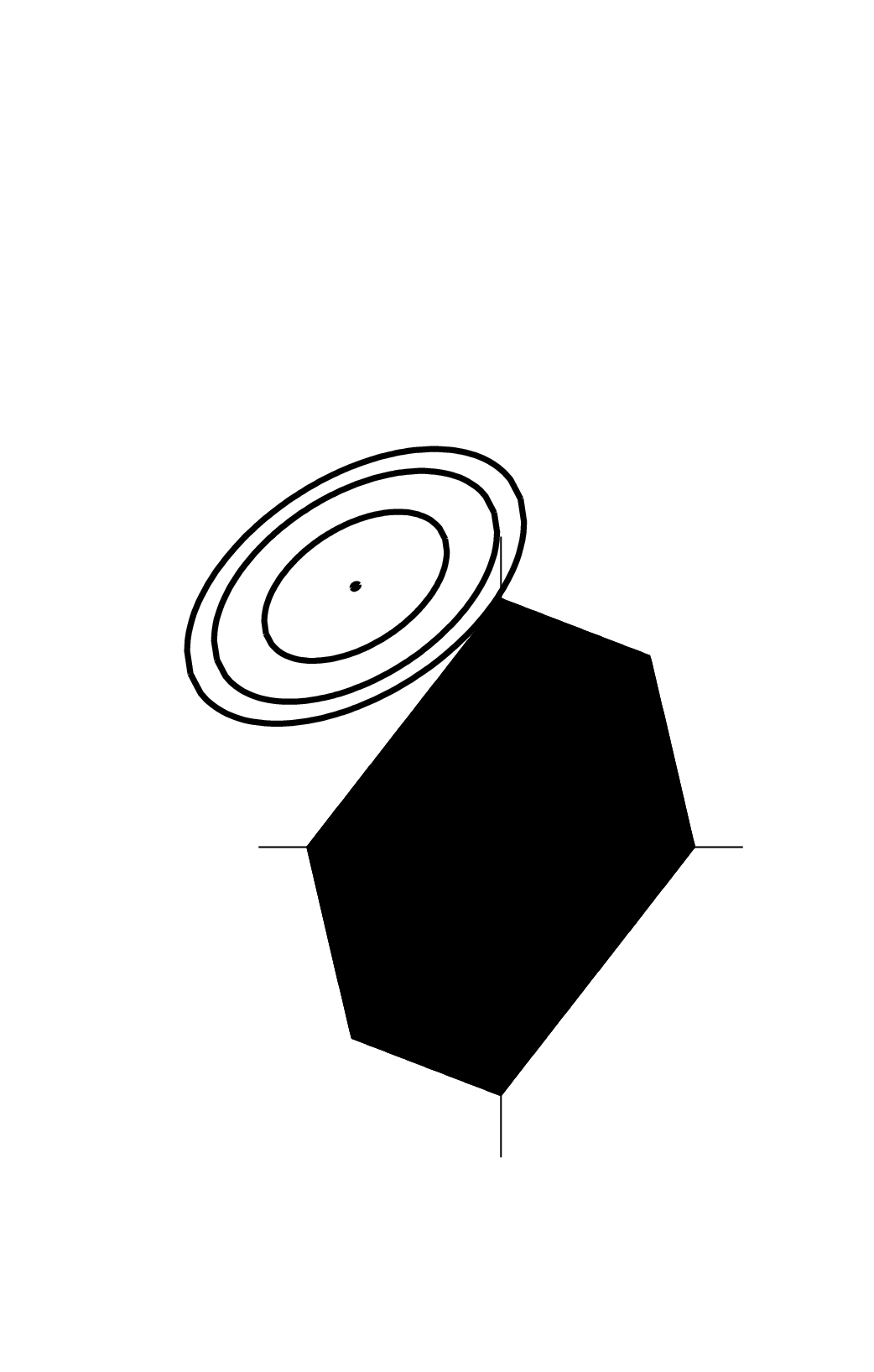}
\label{fig:horses2}
}
 \subfigure[]{
\includegraphics[width=1in, angle=0]{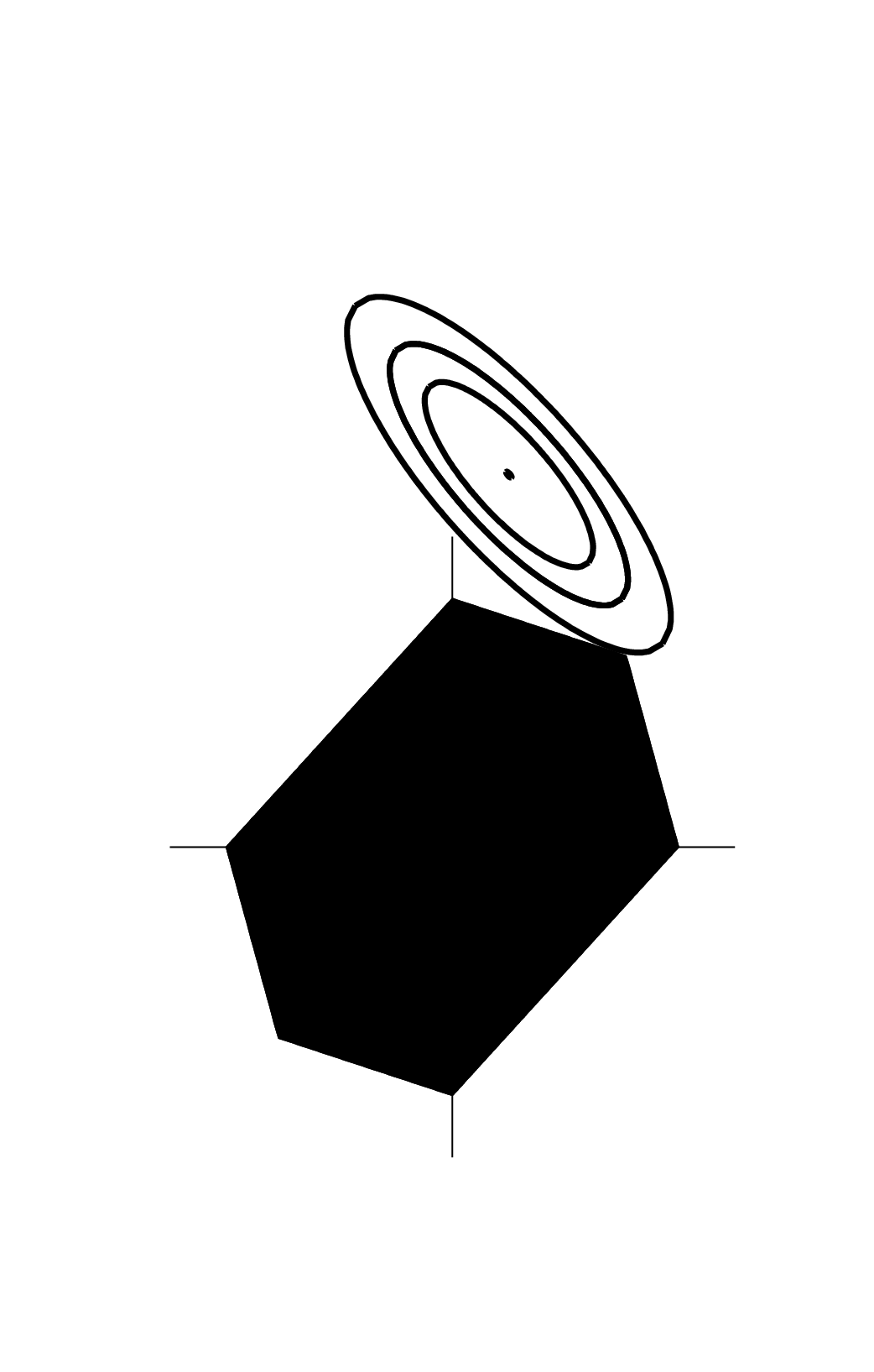}
\label{fig:horses3}
}

\caption{ Graphical representation in the $(\beta_1, \beta_2)$
plane.   HORSES  solutions are the first time the
contours of the sum of squares function hit the hexagonal
constraint region.  \subref{fig:horses2} Contours centered at
OLS estimate with a negative correlation. Solution occurs at $\hat
\beta_1 =0$; \subref{fig:horses3} Contours centered at OLS
estimate with a positive correlation. Solution occurs at $\hat
\beta_1 =\hat \beta_2$.} 
\end{figure}

In a graphical representation in the 
$(\beta_1, \beta_2)$ plane, the solution is the first time the contours of the sum of squares loss function hit the constraint regions.  Figure 2 gives a schematic view.  Figure \ref{fig:horses2} shows the solution
for  HORSES when there is negative correlation between predictors.   HORSES treats them separately
by making $\hat\beta_1 = 0$.  On the other hand, HORSES yields $\hat\beta_1 = \hat\beta_2$ when predictors are positively correlated, as in  Figure    \ref{fig:horses3}.  

The following theorem shows that HORSES has  the  {exact} grouping property. As the correlation between two predictors increases, the predictors are more likely to be grouped together.  Our proof follows closely the proof  of Theorem 1 in \citet{Bondell08} and is hence relegated to an Appendix.

\begin{Thm}

Let $\lambda_1= \lambda \alpha$ and $\lambda_2=\lambda ( 1-
\alpha)$ be the two tuning parameters in the HORSES criterion. Given data
$\big({ y}, { X} \big)$ with centered response ${y}$ and
standardized predictors ${X}=(x_1, \ldots, x_p)^t$, let $\widehat{\beta}
\big(\lambda_1,\lambda_2)$ be the HORSES estimate using the tuning
parameters $\big(\lambda_1,\lambda_2\big)$.   Let $\rho_{kl}
={x}_k^{\rm T} {x}_l$ be the sample correlation between
covariates $x_k$ and $x_l$.

For a given pair of predictors ${x}_k$ and ${x}_l$,
suppose that both $\widehat{\beta}_k (\lambda_1,\lambda_2)$ and
$\widehat{\beta}_l(\lambda_1,\lambda_2)$ are distinct from the
other $\widehat{\beta}_m$. Then there exists $\lambda_0 \ge 0$
such that if $\lambda > \lambda_0$ then
\begin{equation} \nonumber
\widehat{\beta}_k \big(\lambda_1,\lambda_2\big) =
\widehat{\beta}_l \big( \lambda_1,\lambda_2\big), \quad \mbox{for
all  $\alpha \in [d^{-1} ,1]$.}
\end{equation}
Furthermore, it must be that
\begin{equation} \nonumber
\lambda_0 \le  \|{ y}\| \sqrt{2 (1 -\rho_{kl} )} \big/ (1- \alpha).
\end{equation}
\end{Thm}
\medskip

The strength with which the
predictors are grouped is controlled by  $\lambda_2$.  If $\lambda_2$
is increased, any two coefficients are morely likely to be equal. When $x_i$ and $x_j$ are positively correlated, Theorem 1 implies that predictors $i$ and $j$ will be grouped and their coefficient estimates almost identical.  

\section{Related work} \label{sect:review}

This brief review cannot do justice to the many variable selection methods that have been developed. We highlight several of them, especially those that have links to our HORSES procedure.

While variable selection in regression is an increasingly important
problem, it is also very challenging, particularly when there is a large number
of highly correlated predictors. Since the important contribution of the least
absolute shrinkage and selection operator (LASSO) method by \citet{Tibshirani96},  many other methods based on regularized or penalized regression have been proposed for parsimonious
model selection, particularly in high dimensions,
e.g. Elastic Net, Fused LASSO, OSCAR  and Group Pursuit methods \citep{Zou05, Tibshirani05, Bondell08, Shen10}. Briefly, these methods involve
penalization to fit a model to data, resulting in shrinkage of the estimators.
Many methods have focused on addressing various possible shortcomings
of the LASSO method, for instance when there is dependence or collinearity
between predictors.



In the LASSO, a bound is imposed on the sum of  the absolute values of the
coefficients:
\begin{eqnarray*}
\hat \beta &=& \argmin_{\beta} \|y  - \sum_{j=1}^p \beta_j x_j \|^2
\mbox { subject to} \sum_{j=1}^p | \beta_j|  \le  t,
\end{eqnarray*}
where $y = (y_1, \dots, y_n)$ and $x_j = (x_{1j}, \dots, x_{nj})$.

The LASSO method is a shrinkage method like ridge regression \citep{Hoerl70}, with automatic variable selection. Due to the nature
of the $L_1$ penalty term, 
LASSO shrinks each coefficient   and selects variables simultaneously.   However, a major
drawback of  LASSO is that if there exists collinearity among a subset of the
predictors, it usually only selects one to represent the  entire collinear  group. Furthermore, LASSO cannot select more than $n$ variables when $p >n$.

One possible approach is to cluster predictors based on the correlation structure and to use averages of the predictors in each cluster as new predictors.  \citet{Park07} used this approach  for gene expression data analysis and introduce  the concept of a {\it super gene}. However, it is sometimes desirable to keep all relevant predictors separate while achieving better predictive performance, rather than to use an average of the predictors. The hierarchical clustering used in \citet{Park07} for grouping does not account for the correlation structure of  the predictors. 

Other penalized regression methods have also been proposed for grouped
predictors \citep{Bondell08,   Tibshirani05, Zou05, Shen10}. All these methods except Group Pursuit work by introducing a new penalty term in
addition to the $L_1$ penalty term of  LASSO to account for
correlation structure. For example, based on the fact that ridge
regression tends to shrink the correlated predictors toward each
other, Elastic Net \citep{Zou05} uses a linear
combination of ridge and LASSO penalties for group predictor selection
and can be computed by solving the following constrained least squares
optimization problem,
\begin{eqnarray*}
\hat \beta &=& \argmin_{\beta} ||y  - \sum_{j=1}^p \beta_j x_j ||^2
\mbox { subject to} \quad
\alpha \sum_{j=1}^p | \beta_j| + (1-\alpha) \sum_{j =1}^p  \beta_j^2  \le t.
\end{eqnarray*}
The second term forces highly correlated predictors to be averaged
while the first term leads to a sparse solution of these averaged
predictors.

\citet{Bondell08} proposed OSCAR (Octagonal Shrinkage and
Clustering Algorithm for Regression), which is defined by
\begin{eqnarray*}
\hat \beta &=& \argmin_{\beta} ||y  - \sum_{j=1}^p \beta_j x_j||^2  \mbox { subject to}  \quad
\sum_{j=1}^p | \beta_j| + c \sum_{j < k}^p \max\{| \beta_j |, |\beta_k| \}\le t.
\end{eqnarray*}
By using a pairwise $L_{\infty}$ norm as the second penalty term,
OSCAR encourages equality of coefficients. The constraint region for the OSCAR procedure is represented by an octagon (see Figure \ref{fig:region1}\subref{fig:oscar1}). Unlike the hexagonal shape of the HORSES procedure, the octagonal shape of the constraint region allows for grouping of negatively as well as positively correlated predictors. While this is not necessarily an undesirable property, there may be instances when a separation of positively and negatively correlated predictors is preferred.

Unlike Elastic Net and OSCAR, Fused
  LASSO \citep{Tibshirani05} was introduced to account for {\it
  spatial} correlation  of predictors. A key assumption in Fused LASSO
is that the predictors have a certain type of ordering. Fused LASSO
solves
\begin{eqnarray*}
\hat \beta &=& \argmin_{\beta} ||y  - \sum_{j=1}^p \beta_j x_j||^2  \mbox { subject to} \quad
\sum_{j=1}^p | \beta_j|  \le t_1 \mbox{ and } \sum_{j =2}^p |
\beta_j - \beta_{j-1}| \le t_2.
\end{eqnarray*}
The second constraint, called a {\it fusion penalty},  encourages sparsity in the differences of
coefficients. The method can theoretically
be extended to  multivariate data, although with a corresponding
increase in computational requirements.

Note that the Fused LASSO signal approximator (FLSA) in \citet{Friedman07} can be considered as a special case of  HORSES with design matrix $X=I$.  We also want to point out that  our penalty function is a convex combination of the $L_1$  norm of the coefficients and the $L_1$ norm of the pairwise differenc
es of coefficients. Therefore, it is not a straightforward extension of Fused LASSO in which each penalty function is constrained separately. \citet{She10} extended Fused LASSO by considering all possible pairwise differences and called it Clustered LASSO.  However, the constraint region of Clustered LASSO does not have a hexagonal shape. As a result, Clustered LASSO  does not have the {\it exact} grouping property of OSCAR. Consequently, \citet{She10} suggested to use  a data-argumentation modification  such as Elastic Net to achieve exact grouping.

Finally, the Group Pursuit method of \citet{Shen10} is a kind
of supervised clustering. With a regularization parameter $t$
and a threshold parameter $\lambda_2$, they define
$$
G(z) = \left\{ \begin{array}{rl} \lambda_2 & \mbox{  if  $|z| > \lambda_2$} \\
  |z| &\mbox{ otherwise,} \end{array} \right.
  $$
and estimate $\beta$ using
\begin{eqnarray*}
\hat \beta &=& \argmin_{\beta}  \| y  - \sum_{j=1}^p \beta_j x_j \|^2   \mbox { subject to} \quad
\sum_{j  < k }^p G ( \beta_j - \beta_{k})  \le t.
\end{eqnarray*}

HORSES is a hybrid of  the Group Pursuit and Fused LASSO  methods and addresses some limitations of the various methods described above. For example, OSCAR cannot handle the high-dimensional data while Elastic Net does not have the exact grouping property.

\section{Computation and Tuning} \label{sect:compute}

A crucial component of any variable selection procedure is an efficient algorithm for its implementation. 
In this Section we describe how we developed such an algorithm for the HORSES procedure. The Matlab code for this algorithm is available upon request. We also discuss here the choice of optimal tuning parameters for the algorithm.

\subsection{Computation}

Solving the equations for the HORSES procedure (\ref{eqn:horses})
is equivalent to solving its Lagrangian counterpart
\begin{equation} \label{lagr_obj}
f(\beta) = \frac{1}{2} \|y - \sum_{j=1}^p \beta_j x_j \|^2 + \lambda_1 \sum_{j=1}^p |\beta_j| 
+ \lambda_2 \sum_{j<k} |\beta_j - \beta_k|,
\end{equation}
where $\lambda_1 = \lambda \alpha$ and $\lambda_2 = \lambda(1-\alpha)$ with $\lambda>0$.

To solve (\ref{lagr_obj}) to obtain estimates for the HORSES procedure, we modify the pathwise coordinate descent algorithm of \citet{Friedman07}. The pathwise coordinate descent algorithm is an adaptation of the coordinate-wise descent algorithm for solving the 2-dimensional Fused LASSO problem with a non-separable penalty (objective) function. Our extension involves modifying the pathwise coordinate descent algorithm to solve the regression problem with a fusion penalty. As shown in \citet{Friedman07}, the proposed algorithm is much faster than a general quadratic program solver. Furthermore, it allows the HORSES procedure to run in situations where $p > n$.

Our modified pathwise  coordinate descent algorithm has two steps, the descent and the fusion steps. 
In the descent step, we run an  ordinary   coordinate-wise descent  procedure 
to sequentially update each parameter $\beta_k$ given the others. The fusion step
is considered when the descent step fails to improve the objective function. 
In the fusion step, we add an equality constraint on pairs of $\beta_k$s 
to take into account potential fusions and do the descent step along with the constraint.
In other words, the fusion step moves given pairs of parameters together under equality 
constraints to improve the objective function. The details of the algorithm are as follows: 

\begin{itemize} 

\item Descent step:

The derivative of (\ref{lagr_obj}) with respect to $\beta_k$ given $\beta_j=\tilde{\beta}_j$, $j \neq k$, is 
\begin{eqnarray}
\frac{\partial f(\beta)}{\partial \beta_k} &=& x_k^T x_k \beta_k - \big(y - \sum_{j \neq k } \tilde{\beta}_j x_j \big)^T x_k
\nonumber \\
&& \quad + \lambda_1 sgn(\beta_k) + \lambda_2 \sum_{j=1}^{k-1} sgn(\tilde{\beta}_j - \beta_k) + \lambda_2 \sum_{j=k+1}^{p} sgn(\beta_k - 
\tilde{\beta}_j ), \label{eqn:deriv_obj}
\end{eqnarray}
where the $\tilde{\beta}_j$'s are current estimates of the $\beta_j$'s and  
$sgn(x)$ is a subgradient of $|x|$. 
The derivative (\ref{eqn:deriv_obj}) is piecewise linear in $\beta_k$ with 
breaks at $\{ 0, \tilde{\beta}_j, j \neq k\}$ unless $\beta_k \notin \{ 0, \tilde{\beta}_j, j \neq k\}$.

\begin{itemize}

\item  If there exists a solution to $\big( \partial f(\beta) \big/ \partial \beta_k \big) = 0$, 
we can find an interval $(c_1, c_2)$ which contains it, and further show that the solution is  
\begin{eqnarray} 
\tilde{\beta}_k &=& sgn\bigg\{ \tilde{y}^T x_k  - \lambda_2 (\sum_{j<k} s_{jk}  + \sum_{j>k} s_{kj}) \bigg\} \nonumber\\
 &&  \qquad \times \frac{ \Big( \big| \tilde{y}^T x_k  - \lambda_2 (\sum_{j<k} s_{jk}  + \sum_{j>k} s_{kj}) \big| - \lambda_1 \Big)_{+} }{x_k^T x_k}, \nonumber 
\end{eqnarray}
where
$\tilde{y} = y - \sum_{j \neq k } \tilde{\beta}_j x_j$, and 
$s_{jk} = sgn(\tilde{\beta}_j - \frac{c_1+c_2}{2})$. 

\item  If there is no solution to
$\big( \partial f(\beta) \big/ \partial \beta_k \big) = 0$, we let 
\begin{equation} \nonumber 
\tilde{\beta}_k = 
\left\{
\begin{array}{ll}
\tilde{\beta}_l &\quad \mbox{if }~~ f(\tilde{\beta}_l) = \min \big\{ f(0), f(\tilde{\beta}_j ), \mbox{ for }j \neq k \big\} \\
0 &\quad \mbox{if }~~ f(0) \le f(\tilde{\beta}_j ), \mbox{ for every }j \neq k. 
\end{array}
\right.
\end{equation}

\end{itemize}

\item Fusion step:

If the descent step fails to improve the objective function $f(\beta)$, we consider the fusion of pairs of $\beta_k$s. 
For every single pair $(k,l), l \neq k$, we consider the equality constraint $\beta_k = \beta_l = \gamma$ and try
a descent move in $\gamma$. The derivative of (\ref{lagr_obj}) with respect to $\gamma$ becomes 
\begin{equation} \nonumber
\begin{array}{lll}
\frac{\partial f(\beta)}{\partial \gamma} &=& (x_k^T x_k + x_l^T x_l) \gamma - \tilde{y}^T (x_k + x_l)+ 2 \lambda_1 sgn(\gamma)\\
& & \qquad  + 2 \lambda_2 \sum_{j< k,l} sgn(\tilde{\beta}_j - \gamma) + 2\lambda_2 \sum_{j>k,l} sgn(\gamma - \tilde{\beta}_j ),
\end{array}
\end{equation}
where $\tilde{y} = y - \sum_{j \neq k,l } \tilde{\beta}_j x_j$. { If the optimal value of $\gamma$ obtained from the descent step improves the objective function,
we accept the move $\beta_k = \beta_l = \gamma$.}

\end{itemize} 

\subsection{Choice of Tuning Parameters}

Estimation of the tuning parameters $\alpha$ and $t$ used in the algorithm above is very
important for its successful implementation, as it is for the other methods of penalized regression. Several methods have been proposed in
the literature, and any of these can be used to tune the parameters of the HORSES procedure. 
$K$-fold cross-validation (CV) randomly divides the data into $K$ roughly
equally sized and disjoint subsets $D_k$, $k=1,\ldots,K$; 
$\bigcup_{k=1}^K D_k =\{1,2,\ldots,n \}$. The CV error is 
defined by 
\begin{equation} \nonumber 
{\rm CV}(\alpha, t) = \sum_{k=1}^K \sum_{i \in D_k } 
\left( y_i - \sum_{j=1}^p \widehat{\beta}_j^{(-k)}(\alpha,t) x_{ij} \right)^2,
\end{equation} 
where $\widehat{\beta}_j^{(-k)}(\alpha,t)$ is the estimate of $\beta_j$ for a given $\alpha$ and $t$ 
using the data set without $D_k$.

{Generalized cross-validation (GCV) and
Bayesian information criterion (BIC) \citep{Tibshirani96, Tibshirani05,  Zou07}  are other popular methods. These are 
defined by}
\begin{eqnarray*} \nonumber
{\rm GCV} (\alpha,t) &=& \frac{{\rm RSS}(\alpha,t)}{ n - {\rm df}},\\
 {\rm BIC} (\alpha, t) &=&  n \times \log \big({\rm RSS}(\alpha,t)
\big) + \log n \times {\rm df}
\end{eqnarray*}
where $\widehat{\beta}_j\big(\alpha,t \big)$ is the estimate of
$\beta_j$ for a given $\alpha$ and $t$, ${\rm df}$ is the degrees of freedom and
 \begin{equation} \nonumber
{\rm RSS} ( \alpha, t ) = \sum_{i=1}^n \left( y_i - \sum_{j=1}^p
\widehat{\beta}_j (\alpha,t)x_{ij} \right)^2.
\end{equation}
Here, the degrees of freedom is a measure of model complexity. To apply
these methods, one must estimate the degrees of freedom \citep{Efron04}. 
Following \citet{Tibshirani05} for Fused LASSO, we use the number of distinct groups of non-zero regression coefficients as an estimate of the degrees of freedom. 

\section{Simulations} \label{sect:simstudy}

We numerically compare the performance of HORSES and several other
penalized methods: ridge
regression, LASSO, Elastic Net, and OSCAR. We do this by generating data based on six models that differ on the number of data points $n$, number of predictors $p$, the correlation structure $\Sigma$ and the true values of the coefficients $\beta$. 
The parameters for these six models are given in Table \ref{table:models}.

\begin{table}
\begin{tabular}{|cccccl|} \hline
Model & $n$ & $p$ & $\Sigma_{i,j}$ & $\sigma$ & $\beta$ \\ \hline
1 & 20 & 8 & $0.7^{|i-j|}$ & 3 &  $(3,2,1.5,0,0,0,0,0)^T$ \\
2 & 20 & 8 & $0.7^{|i-j|}$& 3 & $(3,0,0,1.5,0,0,0,0,2)^T$ \\
3 & 20 & 8 & $0.7^{|i-j|}$ & 3 & $(0.85,0.85,0.85,0.85.0.85,0.85,0.85,0.85)^T$ \\
4& 100 & 40 & 0.5 & 15 &  $(\underbrace{0,\ldots,0}_{10},\underbrace{2,\ldots,2}_{10},
 \underbrace{0,\ldots,0}_{10},\underbrace{2,\ldots,2}_{10})^T$ \\
 5 & 50 & 40 & 0.5 & 15 & $(\underbrace{3,\ldots,3}_{15},\underbrace{0,\ldots,0}_{25})^T$ \\
 6 & 50 & 100 & $0.7^{|i-j|}$ & 3 & see text \\ \hline
\end{tabular} 
\caption{Parameters for the models used in the simulation study.} \label{table:models}
\end{table}

The first five models are very similar to those in \citet{Zou05} and \citet{Bondell08}.  
Specifically, the data are generated from the model
\begin{equation} \nonumber
 {y} = {\rm  X} {\bf \beta} + \epsilon,
\end{equation}
where $\epsilon \sim N\big(0,\sigma^2\big)$.   
For models 1-4, we generate predictors $x_i =(x_{i1}, \ldots, x_{ip})^{t}$ from a multivariate normal distribution with mean 0 and covariance $\Sigma$ where $\Sigma_{j,j} =1$ for $j=1, \ldots,p$. 

For model 5, the predictors are generated as follows:
\begin{eqnarray*}
&&x_i = Z_1 + \eta_i^x, Z_1 \sim N(0,1), \quad i \in G_1 =\{1, \ldots, 5\}\\
&&x_i = Z_2 + \eta_i^x, Z_2 \sim N(0,1), \quad i \in G_2 =\{6, \ldots, 10\}\\
&&x_i = Z_3 + \eta_i^x, Z_3 \sim N(0,1), \quad i \in G_3=\{11, \ldots, 15\}\\
&&x_i \sim N(0,1), i=16,\ldots, 40.
\end{eqnarray*}
where $\eta_i^x \sim N(0,0.16), i=1,\ldots, 15$. Then  Corr$(x_i,x_j) \approx 0.85$ for $i, j \in G_k$ for $k=1,2,3$. 

For model 6, we consider the scenario where  $p>n$. We choose $p=100$  because this is  the maximum number of predictors  that can be handled by the quadratic programming used in OSCAR.  
The vector of coefficients $\beta$ for model 6 is given by
\begin{equation} \nonumber
\beta =
(\underbrace{3,\ldots,3}_{5},\underbrace{0,\ldots,0}_{10}, \underbrace{2,\ldots,2}_{5},\underbrace{0,\ldots,0}_{10}, \underbrace{-1.5,\ldots,-1.5}_{5},\underbrace{0,\ldots,0}_{10}, \underbrace{1,\ldots,1}_{5},\underbrace{0,\ldots,0}_{50})^T
\end{equation}

We generate 100 data
sets of size $2n$ for each of the 6 models. In each data set,
the final model is estimated as follows: (i) For each
$(\alpha,t)$, we use  the first $n$ observations as a training set to estimate the model and use  the other $n$ observations as a validation set to compute the prediction error ${\rm PE}(\alpha,t)$; (ii) We set the tuning parameters to be the values
$(\alpha^*,t^*)$ that minimize the prediction error ${\rm
PE}(\alpha,t)$; (iii) The final model is estimated using the training set with $(\alpha,t)=(\alpha^*,t^*)$.

\begin{table}[b]
\begin{center}
\caption{The number of groups in each model used in the simulation study.}
\begin{tabular}{lcccccc}\\\hline
Model & 1 & 2 & 3 & 4 & 5 & 6 \\ \hline
Number of groups & 3 & 3 & 1 & 1 & 3 & 4 \\\hline
\end{tabular}
\end{center}
\end{table}

We compare the mean square error (MSE) and the
model complexity of the five penalized methods. The MSE is
calculated as in \citet{Tibshirani96} via
\begin{equation} \nonumber
 {\rm MSE} = (\hat{\beta} - \beta)^T {\rm V}
(\hat{\beta} - \beta),
\end{equation}
where  ${\rm V}$ is the population covariance matrix for $X$. The model complexity is measured by the number of groups. Based on the coefficient values and correlation structure, Table 1 shows the true number of groups for each of the six scenarios. Note that the true number of groups is not always the same as  the degrees of freedom. For example, we note that the true number of groups in model 5 is three based on the correlation structure although all nonzero coefficients have the same value. On the other hand, model 4 assumes a compound symmetric covariance structure, therefore the number of groups only depends on the coefficient values.  Hence, the order of the coefficients  does not matter and we can consider model 4 as having only one group of non-zero coefficients. We take the model complexity of model 6 to be four, based on the coefficient values. However, it is possible that some of the zero coefficients might be included as signals because of strong correlations and relatively small differences in coefficient values in this case. For example, the correlation between $\beta_{50}=1$ and $\beta_{51}=0$ is 0.7. Therefore it is possible that the true model complexity  in this case may be bigger than four. 

The simulation results are summarized in Table 2. The HORSES procedure
reports the smallest {\rm df}s  except for models 1 and 6. In both scenarios, the differences of {\rm df} between the least complex model and HORSES  
is marginal (4 vs 5 in model 1 and 30 vs 33.5 in model 6). The HORSES procedure
is also very competitive in the MSE comparison. Its MSE is the
smallest in models 2-4 and 6 and the second or third smallest in models 1 and 5.

It is interesting to observe that HORSES is the best in model 2, but
third in model 1 although the differences in  MSE and {\rm df} of Elastic Net and HORSES
in model 1 are minor. The values of the parameters are the same in
both scenarios, but  variables with similar coefficients are highly
correlated in model 1, while these variables have little correlation
with each other in model 2. Hence we can consider the grouping of predictors  as
mainly determined by coefficient values in model 2 while in model 1, the correlation 
structure may have an  important role in the grouping. This can be confirmed by comparing the median 
MSEs of each method  in the two models 1 and 2. As expected, the median MSE in model 1 is always smaller than the median MSE in model 2. The difference in the median MSEs can be interpreted  as the gain achieved by using the correlation structure when grouping. Because of the
explicit form of the fusion penalty in HORSES,  our procedure seems
to give more weight to differences among the coefficient values while
still accounting for correlations. As a result,
HORSES effectively groups in model 2.  Not surprisingly, the
HORSES procedure is much more successful than the other procedures in finding
the correct model in model 3, where it
may give higher weight to the fusion penalty ($\alpha$
close to $1$). It also has the smallest MSE among the methods. In this case, the true model is not sparse and the LASSO and Elastic Net methods fail.  HORSES outperforms the other methods again in
model 4. Since the model assumes the compound symmetric covariance
structure, the grouping is solely based on the coefficient values.
Because of the fusion penalty, the HORSES procedure is very effective
in grouping  and produces 3.5 as the median {\rm df} while the second
smallest {\rm df} is 15 with OSCAR. In model 5, HORSES has the second
smallest median MSE (=46.1) with Elastic Net's median MSE smallest at
40.7. However, HORSES chooses the least complex model and shows better
grouping compared to Elastic Net. Model 6 considers a  large $p$ and  small $n$ case.  The HORSES procedure reports the smallest MSE while the Elastic Net chooses the least complex model. However we notice that all methods report at least 30 as the {\rm df}. This might be due to the fact that the true model complexity in this case is not clear, as we point out above.    In summary, the HORSES procedure outperforms the other methods in choosing the least complex model and
attaining the best grouping, while also providing competitive results in terms of MSE. 

\begin{table}[ht]
\begin{center}
\caption{MSE and model complexity.}
\begin{tabular}{llcccccc}\\\hline
 Case & Method & MSE & MSE  & MSE &DF &DF &DF \\
& &Med. & 10th perc. & 90th perc. &Med. & 10th perc. & 90th perc.\\\hline
  {\bf C1} & Ridge &  2.31 & 0.98 &  4.25&8 &8 &8 \\
    & LASSO & 1.92 & 0.68 & 4.02 & 5& 3&8 \\
  & Elastic Net & 1.64& 0.49 & 3.26& 5&3 &7.5 \\
  & OSCAR & 1.68 & 0.52 & 3.34 &4 &2 &7 \\
  & HORSES & 1.85 &  0.74 & 4.40&5&3 &8 \\\hline
{\bf C2} & Ridge & 2.94  & 1.36 & 4.63 &8 &8 &8 \\
    & LASSO & 2.72 & 0.98 & 5.50&5 &3.5 &8 \\
  & Elastic Net & 2.59 & 0.95 & 5.45 &6 &4 &8 \\
  & OSCAR & 2.51  & 0.96  & 5.06&5 &3 &8 \\
  & HORSES & 2.21 & 1.03 & 4.70 &5 &2 &8 \\ \hline
  {\bf C3} & Ridge &  1.48 & 0.56 & 3.39 &8 &8 &8 \\
    & LASSO & 2.94 & 1.39 & 5.34&6 &4 &8 \\
  & Elastic Net & 2.24& 1.02& 4.05&7 &5 & 8\\
  & OSCAR & 1.44 & 0.51& 3.61&5 &2 &7 \\
  & HORSES & 0.50 & 0.02 & 2.32 &2 &1 &5.5 \\ \hline
  {\bf C4} & Ridge &  27.4 & 21.2 &  36.3&40 &40 &40 \\
    & LASSO & 45.4&32 &56.4 & 21&16 &25 \\
  & Elastic Net & 34.4& 24 & 45.3&25 &21 &28 \\
  & OSCAR & 25.9 & 19.1 & 38.1&15 &5 &19 \\
  & HORSES & 21.2&19.3 & 33.0 &3.5 &1 &19.5 \\\hline
   {\bf C5} & Ridge &  70.2 & 41.8 &  103.6&40 &40 &40 \\
    & LASSO & 64.7&27.6 &116.5 &12 &9 &18 \\
  & Elastic Net & 40.7& 17.3 & 94.2&17 &13 &25 \\
  & OSCAR & 51.8 & 14.8 & 96.3&12 &9 &18 \\
  & HORSES & 46.1 &18.1 & 92.8 &11 &5.5 &19.5  \\\hline
    {\bf C6} &	Ridge	&	27.71 	&	19.53 	&	38.53 	&	100 	&	100 	&	100 	\\
&	LASSO	&	13.36 	&	7.89 	&	20.18 	&	31	&	24 	&	39.1 	\\
&	Elastic Net	&	13.57 	&	8.49 	&	25.33 	&	30 	&	23.9 	&	37	\\
&	OSCAR	&	13.16 	&	8.56 	&	19.16 	&	50.00 	&	35.9 	&	83.7 	\\
&	HORSES	&	12.20 	&	7.11 	&	22.02 	&	33.5 	&	24 	&	66.3 	\\\hline
 \end{tabular}
\end{center}
\end{table}

\section{Data Analysis} \label{sect:analysis}

 \subsection{Cookie dough data} 

\begin{figure}[t]
\begin{center}
\includegraphics[scale=0.5,angle=0]{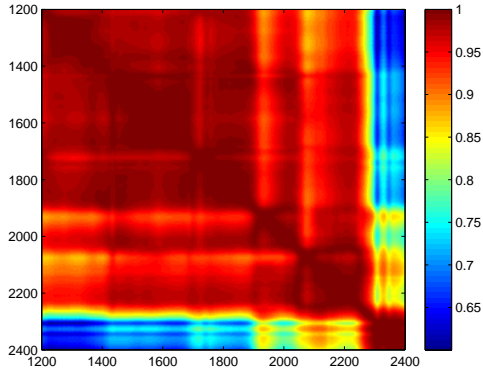}
\end{center}
\caption{Graphical representation of the correlation matrix of the 300 wavelengths of the cookie dough data}
\label{fig:doughcorr}
\end{figure}
In this case study, we consider the cookie dough dataset from \cite{Osborne1984}, which was also analyzed by \cite{Brown2011}, \cite{Griffin2007}, \citet{Caron08},
 and \cite{Hans2011}.  \cite{Brown2011} consider four components as response variables:  percentage of fat, sucrose, flour and water associated with each dough piece.  Following \citet{Hans2011},  we attempt to predict only the flour content of cookies with the 300 NIR reflectance measurements at equally spaced wavelengths between 1200 and 2400 nm as predictors (out of the 700 in the full data set). {Also following \citet{Hans2011} we remove the 23rd and 61st observations as outliers. Then} we split the dataset randomly into a training set with 39 observations and 
a test set with 31 observations.   Figure \ref{fig:doughcorr} shows the correlations between NIR reflectance measurements based on all observations. There are very strong correlations between any pair of predictors in the range of 1200-2200 and 2200-2400. Note however that strong correlations do not necessarily imply strong signals in this case since the correlations can be due to the measurement errors.

\begin{figure}[t]
\centering
\subfigure[Elastic Net]{
\includegraphics[scale=0.2,angle=-90]{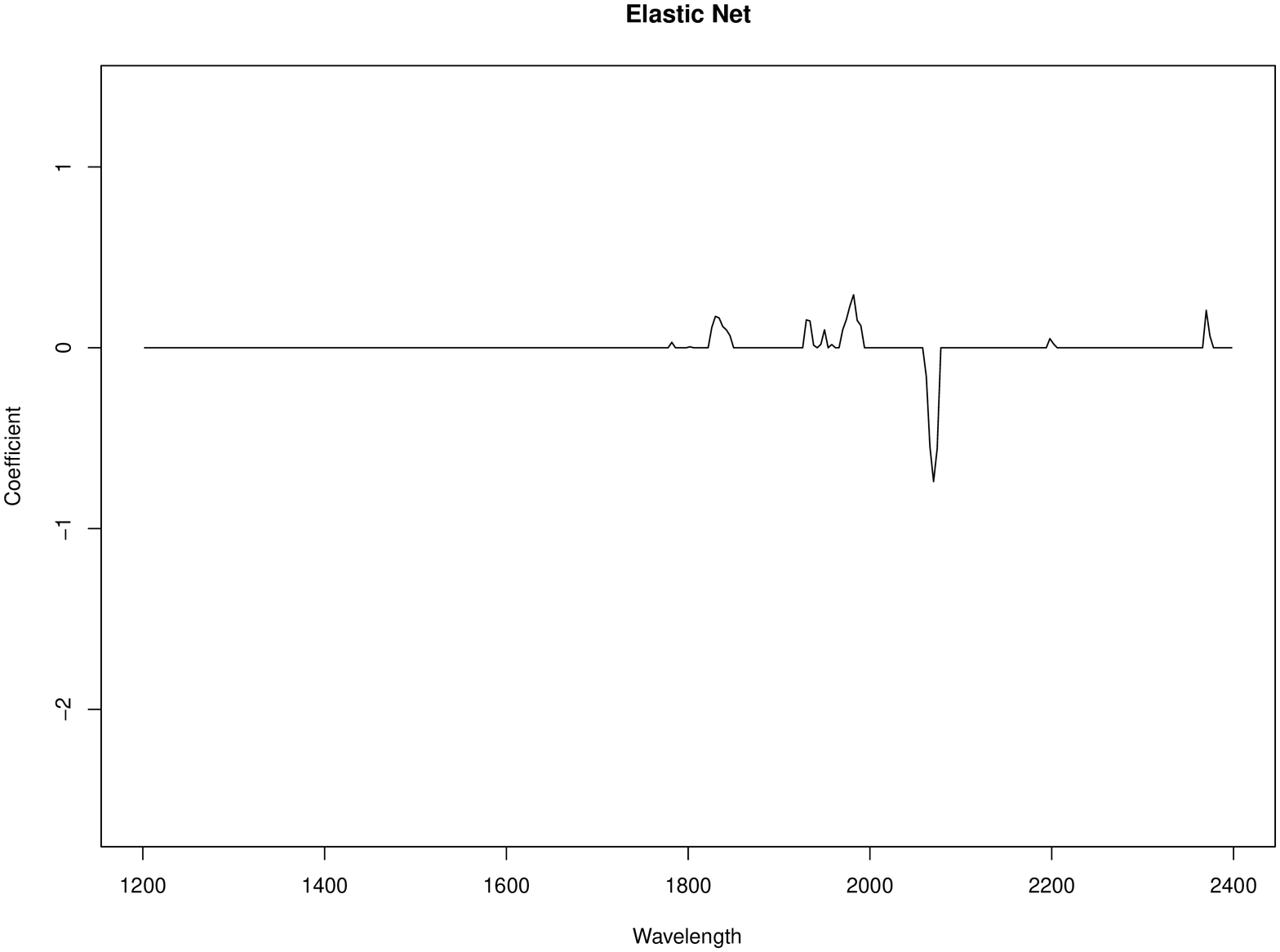}
\label{fig:biscuit_net}
}
 \subfigure[LASSO]{
\includegraphics[scale=0.2, angle=-90]{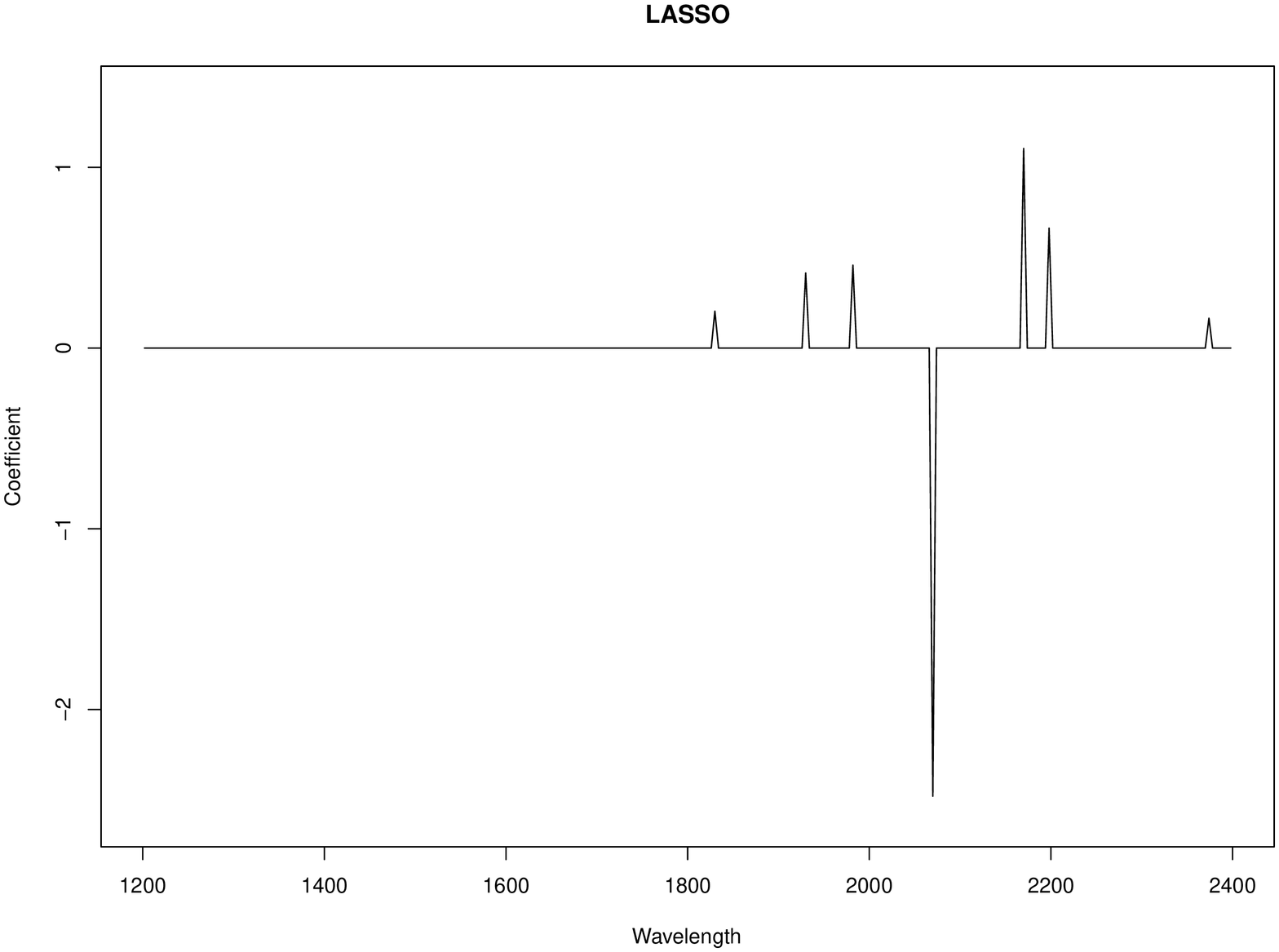}
\label{fig:biscuit_lasso}
}
\subfigure[HORSES]{
\includegraphics[scale=0.2, angle=-90]{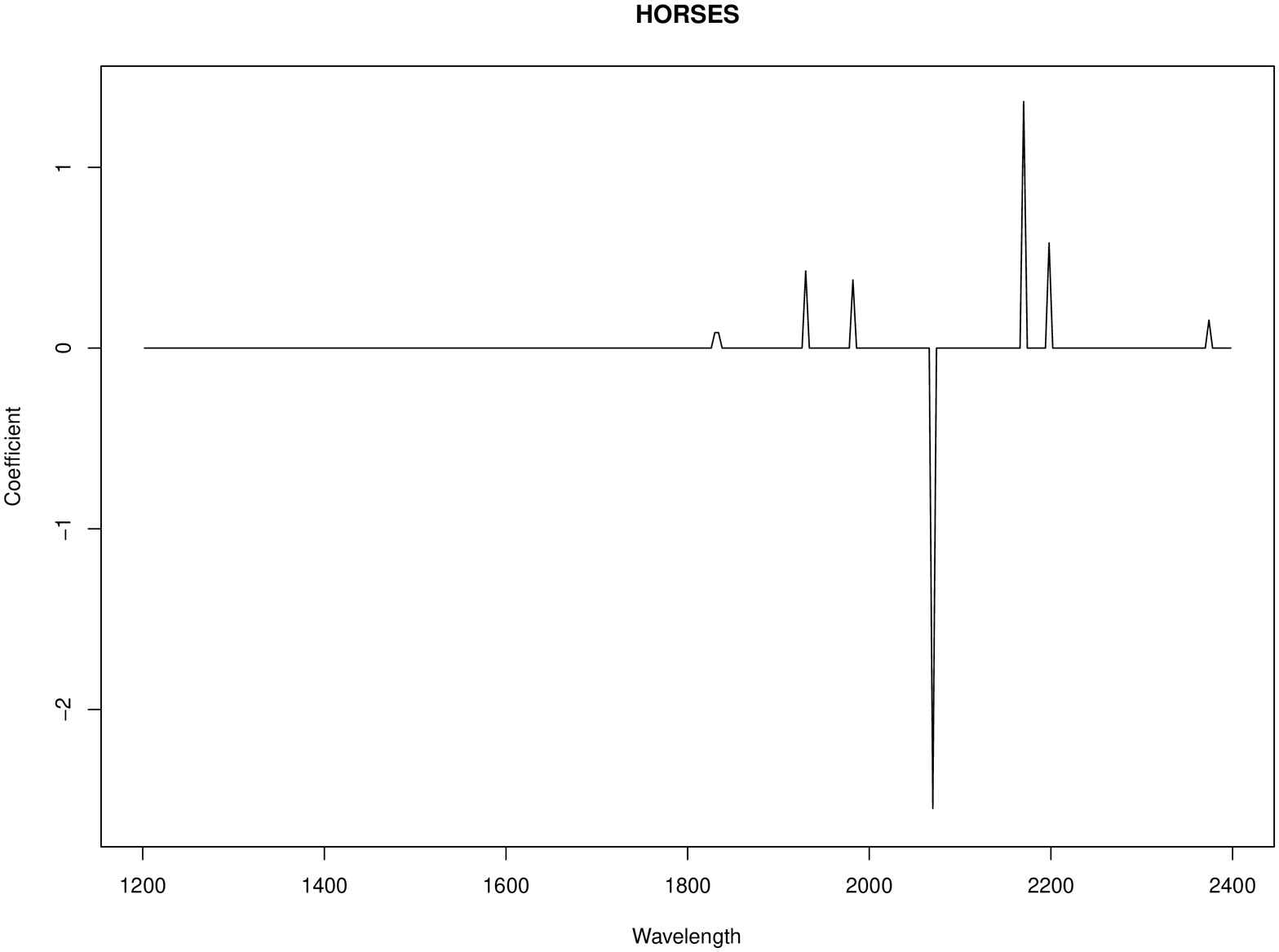}
\label{fig:biscuit_horses} } \caption{Coefficient estimates for the 300 predictors of the cookie dough data}.
\label{fig:biscuit}
\end{figure}

With the training data set,  tuning parameters of HORSES are computed to be $\alpha=0.999$ and $\lambda=0.1622$ (equivalently, $\lambda_1=0.1620$ and  $\lambda_2=0.00016$). Since the $L_1$ penalty dominates the penalty function, we expect  that both HORSES and LASSO will yield very similar results. We compare HORSES, LASSO and Elastic Net via the prediction mean squared error and degrees of freedoms on the test data.  The OSCAR method is not included in the  comparison because  we are not able to apply  it  due to the high dimension of the data.  Table  3 presents the prediction mean squared error and degrees of freedom of each method.  The Elastic Net has the smallest MSE, but the differences  in MSE across the three methods are small. On the other hand, the LASSO and HORSES methods provide parsimonious models with small degrees of freedom.  
The estimated coefficients for the LASSO, Elastic Net and HORSES methods are presented  in Figure  \ref{fig:biscuit}.   Elastic Net produces 11 peaks while both LASSO and HORSES have 7 peaks.  The estimated spikes from  LASSO and HORSES are consistent with the results obtained in \citet{Caron08}.  The main difference between the two methods is at wavelengths 1832 and 1836, where the LASSO estimates are 0.204 and 0 while the HORSES estimates are 0.0853 at both wavelengths. The Elastic Net has peaks at wavelength 1784 and 1804 but the other two methods do not provide a peak at those wavelengths.  We observe a reverse pattern at wavelength 2176.

\begin{table}[ht]
\caption{Biscuit dough data results}
\begin{center}
\begin{tabular}{cccc}\\\hline 
& Elastic Net & HORSES & LASSO\\\hline
 Mean Squared Error & 2.442 & 2.586 & 2.556 \\
 Degrees of Freedom  & 11 & 7 & 7  \\\hline
\end{tabular}

\end{center}
\end{table}

\subsection{Appalachian Mountains Soil Data}
\begin{figure}[t]
\begin{center}
\includegraphics[scale=0.5,angle=0]{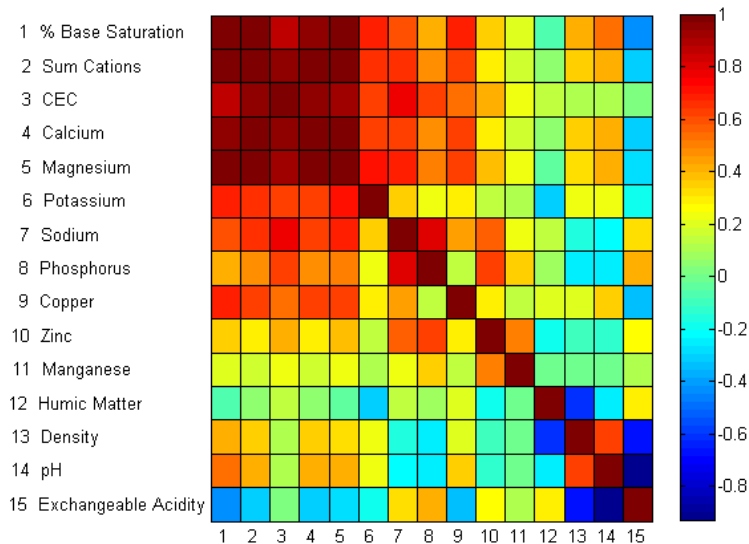}
\end{center}
\caption{Graphical representation of the correlation matrix of the 15 predictors of the Appalachian soil data}
\label{fig:soilcorr}
\end{figure}

Our next example is the Appalachian Mountains Soil Data from \citet{Bondell08}. Figure \ref{fig:soilcorr} shows a graphical representation of the
correlation matrix of  15 soil characteristics computed from
measurements made at twenty 500-$m^2$ plots located in the Appalachian
Mountains of North Carolina. The data were collected as part of a
study on the relationship between rich-cove forest diversity and soil
characteristics. Forest diversity is measured as the number of
different plant species found within each plot. The values in the soil
data set are averages of five equally spaced measurements taken within
each plot and are standardized before the data analysis. These soil
characteristics serve as predictors with forest diversity as the response.

As can be seen from Figure \ref{fig:soilcorr}, there are several
highly correlated predictors. Note that our correlation graphic shows
the signed correlation values and is thus
different from the one in \citet{Bondell08} showing the {\it
  absolute} value of correlation.   
{The first seven covariates are closely related. Specifically they
concern positively charged ions (cations). The predictors named ``calcium'', ``magnesium'',
``potassium'', and ``sodium'' are all measurements of cations of the
corresponding chemical elements, while ``\% Base Saturation'', ``Sum Cations'' and ``CEC''
(cation exchange capacity) are all summaries of cation abundance.
The correlations between these seven covariates fall in the range
(0.360, 0.999). There is a very strong positive correlation 
between percent base saturation and calcium ($r=0.98$), but the correlation
between potassium and sodium  ($r=0.36$) is not quite as high as the
others. Of the remaining eight variables, the strongest negative
correlation is between soil pH and exchangeable acidity
($r=-0.93$). Since both of these are measures of acidity, this appears surprising. However, exchangeable acidity measures only
a subset of the acidic ions measured in pH, this subset being of more
significance only at low pH values.}

{Note that because ``Sum Cations'' is the sum of the other four
cation measurements the 
design matrix for these predictors is not full rank.}

\begin{table}[t]
\begin{center}
\caption{Results of analyzing the Appalachian soil data using OSCAR and HORSES,
and two different methods for choosing the tuning parameters.} 
 \begin{tabular}{l|ccccc}\hline
  Variable & OSCAR  & OSCAR & HORSES & HORSES \\
  & (5-fold CV) & (GCV) & (5-fold CV) & (GCV) \\\hline
  \% Base saturation    &   0   &   -0.073  &   0   &   {-0.1839}    \\
  Sum cations   &   {-0.178}    &   {-0.174}    &   {-0.1795}   &   {-0.1839}    \\
  CEC   &   {-0.178}    &{-0.174}   &   {-0.1795}   &   {-0.1839}    \\
  Calcium   &   {-0.178}    &   {-0.174}    &   {-0.1795}   &   {-0.1839}    \\
  Magnesium     &   0   &   0   &   0   &   0    \\
  Potassium     &   {-0.178}    &   {-0.174}    &   {-0.1795}   &   {-0.1839}    \\
  Sodium    &   0   &   0   &   0   &   0    \\
  Phosphorus    &   0.091   &   {0.119} &   0.0803  &   {0.2319}     \\
  Copper    &   0.237   &   {0.274} &   0.2532  &   {0.3936}     \\
  Zinc  &   0   &   0   &   0   &   -0.0943  \\
  Manganese     &   0.267   &   {0.274} &   0.2709  &   {0.3189}     \\
  Humic matter  &   -0.541  &   -0.558  &   -0.5539 &   -0.6334  \\
  Density   &   0   &   0   &   0   &   0    \\
  pH    &   0.145   &   {0.174} &   0.1276  &   {0.2319}     \\
  Exchangeable acidity  &   0   &   0   &   0   &   0.0185  \\\hline\hline
Degrees of Freedom & 6 & 5 & 6 & 7 \\\hline
 \end{tabular}
 \end{center}
\end{table}

We analyze the data with the HORSES and OSCAR procedures and report
the results in Table 4. Although  OSCAR and HORSES use the same definition of df, the
 OSCAR  procedure groups predictors based on the {\it absolute} values of the coefficients.  Therefore the number of groups is not the same as the df in  OSCAR. The results for LASSO using 5-fold
cross-validation and GCV can be found in \citet{Bondell08}. The 5-fold
cross-validation OSCAR and HORSES solutions are similar. They
select the exact same variables, but with slightly different coefficient
estimates. Since the sample size is only 20 and the number of
predictors is 15, the 5-fold cross-validation method may not be the
best choice for selecting tuning parameters.  However, using GCV,  OSCAR
and HORSES provide different answers. Compared to
the 5-fold cross-validation solutions, the OSCAR solution has one more predictor (\%
Base saturation) while the HORSES solution has 3 additional
predictors (\% Base saturation, Zinc, Exchangeable acidity). More
interestingly, in the OSCAR solution, \% Base saturation is not in the
group measuring {\it
  abundance of cations}, while pH is.  
  
  On the other hand, the \% Base
saturation variable is included in the {\it abundance of cations} group.
The HORSES solution also produces an additional group of variables
consisting of Phosophorus and pH.

 \section{Conclusion} \label{sect:conclusion}

{We proposed a new group variable selection procedure in regression that produces a
sparse solution and also groups positively correlated variables
together. We developed a modified pathwise coordinate optimization for applying the
procedure to data.  Our algorithm is much faster than a quadratic program solver and can handle cases with $p>n$.

Such a procedure is useful relative to 
other available methods in a number of ways. First, it selects groups
of variables, rather than randomly selecting one variable in the group
as the LASSO method does. Second, it groups positively correlated
rather than both positively and negatively correlated variables. This
can be useful when studying the mechanisms underlying a process, since
the variables within each group behave similarly, and may indicate
that they measure characteristics that affect a system through the
same pathways. Third, the penalty function used ensures that the
positively correlated variables do not need to be spatially
close. This is particularly relevant in applications where spatial contiguity is not the only indicator of functional relation, such as brain imaging or genetics.}

{A simulation study comparing the HORSES procedure with ridge
  regression, LASSO, Elastic Net and OSCAR methods over a variety of
  scenarios showed its superiority in terms of sparsity, effective
  grouping of predictors and MSE.}

It is desirable to achieve a theoretical optimality such as the oracle property of \citet{Fan01}  in high dimensional cases. One possibility is to extend the idea of  the adaptive Elastic Net \citep{Zou09} to the HORSES procedure. Then we may consider the following penalty form:
    
\begin{eqnarray*}
\hat \beta &=& \mbox{argmin}_{\beta} \|y  - \sum_{j=1}^p \beta_j x_j \|^2  \mbox { subject to} \\
&&\alpha \sum_{j=1}^p  \hat w_j | \beta_j| + (1-\alpha)\sum_{j < k} | \beta_j - \beta_k| \le t,
\end{eqnarray*}    
where $\hat w_j$ are the adaptive data-driven weights. 

Investigating theoretical properties of the above estimator will be a topic of future research.



\section{Appendix} \label{sect:appendix}
{\bf Proof of Theorem 1:}

Note that one can write the HORSES optimization problem in the equivalent Lagrangian form 
\begin{equation} \label{eqn:HORSES-obj}
 \argmin_{\beta} \left\{ \|{ y} - \sum_{j=1}^p \beta_j { x}_j \|^2 +
\lambda \left( \alpha\sum_{j=1}^p | \beta_j | +  (1-\alpha) \sum_{j < k} |
\beta_j - \beta_k | \right) \right\}.
\end{equation}

Suppose the covariates $({x}_1,{
x}_2,\ldots,{ x}_p)$ are ordered such that their corresponding
coefficient estimates satisfy
\begin{equation} \nonumber
\widehat{\beta}_1 \le \widehat{\beta}_2\le \cdots \le
\widehat{\beta}_{\rm L}< 0 < \widehat{\beta}_{{\rm L}+1} \cdots
\le \widehat{\beta}_Q
\end{equation}
and $\widehat{\beta}_{Q+1}=\cdots=\widehat{\beta}_p=0$.
Let $\widehat{\theta}_1, \ldots,\widehat{\theta}_G$ denote the $G$
unique nonzero values of the set of $\widehat{\beta}_j$, so that $G
\le Q$. For each $g=1,2,\ldots,G$, let
\begin{equation} \nonumber
\mathcal{G}_g=\{j : \widehat{\beta}_j = \widehat{\theta}_g\}
\end{equation}
denote the set of indices of the covariates whose estimates of regression coefficients are $\widehat{\theta}_g$.  Let also $w_g=|\mathcal{G}_g|$ be the number 
of elements in the set $\mathcal{G}_g$

Suppose that $\widehat{\beta}_k (\lambda_1,\lambda_2) \neq
\widehat{\beta}_l (\lambda_1,\lambda_2)$ and both are non-zero. In addtion, let assume $k \in \mathcal{G}_g$ and $l \in \mathcal{G}_h$ for $h>g$ without loss of generality. The differentiation of the objective function (\ref{eqn:HORSES-obj}) with respect
to $\beta_k$ gives  
\begin{equation} \nonumber
-2  {x}_k^{\rm T} \big( y - \sum_{j=1}^p \widehat{\beta}_j x_j)+ \lambda \kappa_k =0,
\end{equation}
where $u_{+,g}=\sum_{g1<g} w_{g1}$ and $u_{g,+} =\sum_{g<g2} w_{g2}$, and 
\begin{equation} 
\kappa_k =\alpha ~sgn(\widehat{\beta}_k) + \big( 1- \alpha \big) \big( u_{+,g} - u_{g,+} \big).  
\end{equation} 
In the same way, the differentiation of  (\ref{eqn:HORSES-obj}) with respect to $\beta_l$ is 
\begin{equation} \nonumber 
-2  {x}_l^{\rm T} \big( y - \sum_{j=1}^p \widehat{\beta}_j x_j)+ \lambda \Big\{
 \alpha~ sgn(\widehat{\beta}_l) + \big( 1- \alpha\big) \big( u_{+,h} - u_{h,+} \big) \Big\}=0,
\end{equation} 
and we have, by taking their differences,  
\begin{equation} \nonumber
-2 \big( {x}_k^{\rm T} - {x}_l^{\rm T} \big)
\big( y - \sum_{j=1}^p \widehat{\beta}_j x_j \big)+ \lambda \big(
\kappa_k - \kappa_l\big) = 0.
\end{equation}

Since ${X}$ is standardized, $\|{ x}_k^{\rm T} - {
x}_l^{\rm T} \|^2=2 (1- \rho_{kl} )$. This together with the fact
that $\| { y} - { X} \widehat{\beta} \|^2 \le \|{ y}
\|^2$ gives
\begin{equation} \nonumber
|\kappa_k -\kappa_l| \le 2 \lambda^{-1} \|{y}\| \sqrt{2
(1-\rho_{ij})}.
\end{equation}
However, we find that 
\begin{eqnarray} 
\kappa_l - \kappa_k &=&  \alpha \big\{ sgn(\widehat{\beta}_l) - sgn(\widehat{\beta}_k)\big\} \nonumber\\
&& \qquad + \big( 1- \alpha \big) 
\Big\{ \big(u_{+,h}-u_{h,+} \big) - \big( u_{+,g} - u_{g,+} \big) \Big\}, 
\end{eqnarray} 
is always larger than or equal to $2(1-\alpha)$. Thus,  
If $2 \lambda^{-1} \|{y}\| \sqrt{2 (1-\rho_{kl})}<2(1-\alpha)$ - equivalently, $   \|{y}\| \sqrt{2 (1-\rho_{kl})} \big/ (1-\alpha)< \lambda$ -  then
we encounter a contradiction.





\end{document}